\documentclass{article}

    \PassOptionsToPackage{numbers, compress}{natbib}

     \usepackage[preprint]{neurips_2020}

\usepackage[utf8]{inputenc} 
\usepackage[T1]{fontenc}    
\usepackage{hyperref}       
\usepackage{url}            
\usepackage{booktabs}       
\usepackage{amsfonts}       
\usepackage{nicefrac}       
\usepackage{microtype}      
\usepackage{amsmath}
\usepackage{xcolor}
\usepackage{multirow}
\usepackage{graphicx}
\usepackage{subcaption}
\title{{B}etter call {S}urrogates: A hybrid {E}volutionary {A}lgorithm for {H}yperparameter optimization}

\author{%
 Subhodip~Biswas
 \\
  Computer Science, Virginia Tech\\
  Arlington, VA 22203 \\
  \texttt{subhodip@cs.vt.edu} \\
   \And
  Adam D. Cobb \\
  US Army Research Laboratory \\
  Adelphi, MD 20783\\
  \texttt{cobb.derek.adam@gmail.com} \\
  \And
  Andreea Sistrunk \\
  US Army, ERDC-GRL\\
  Alexandria, VA 22315 \\
  \texttt{sistrunk@vt.edu} \\
  \And
  Naren Ramakrishnan \\
  Computer Science, Virginia Tech\\
  Arlington, VA 22203 \\
  \texttt{naren@cs.vt.edu} \\
   \And
  Brian Jalaian \\
  US Army Research Laboratory \\
  Adelphi, MD 20783\\
  \texttt{brian.a.jalaian.civ@mail.mil} \\
}

   \newcommand{\algo}{\textsc{SteaDE}} 
   \newcommand{\botorch}{\textsc{BoTorch}}
   \newcommand{\pysot}{{pySOT}}
   \newcommand{\poap}{{POAP}} 
   \def\u{{\vphantom g}_}
   \usepackage[ruled, vlined]{algorithm2e}

\begin{document}

\maketitle

\begin{abstract}
In this paper, we propose a surrogate-assisted evolutionary algorithm (EA) for hyperparameter optimization of machine learning (ML) models. The proposed \algo~model initially estimates the objective function landscape using Radial Basis Function interpolation, and then transfers the knowledge to an EA technique called Differential Evolution that is used to evolve new solutions guided by a Bayesian optimization framework. We empirically evaluate our model on the hyperparameter optimization problems as a part of the black box optimization challenge at NeurIPS 2020, and demonstrate the improvement brought about by \algo~ over the vanilla EA.
 
\end{abstract}

\section{Introduction}
\label{sec:intro}
The empirical performance of ML models can be enhanced by configuring its hyperparameters. However, manually searching for a good configuration of hyperparameters is nearly impossible due to the exponentially large size of the parameter space. This has led to an increased interest from the research community of recent in devising automated ways of searching for optimal hyperparameter configuration, popularly known as \textit{hyperparameter optimization} (HPO)~\cite{feurer2019hyperparameter}. HPO is a blackbox optimization problem, where the objective function $f$ is usually \textit{expensive} to evaluate, and no other information like gradient, presence of special structures like linearity/concavity is available. The objective in HPO is to seek the optimum of blackbox function $f$, which is usually some error metric quantifying the performance of a ML model, by expending as little computational budget as possible. 

EAs belong to a class of stochastic derivative-free optimization algorithms~\cite{eiben2015introduction}. In spite of being fast, EAs employ random search moves that result in low sample efficiency, i.e., the expected gain in the objective function per unit of function evaluation (FE). Hence EAs are preferred when the FEs are cheap to compute and a high evaluation budget is available.
On the other hand, surrogate-based optimization algorithms have high data efficiency as they approximate the objective function~\cite{surrogate}, but incur high cost of suggesting new points and hence they are used in solving computationally expensive optimization problems~\cite{practicalBO,brochu2010tutorial}. Radial Basis Functions (RBFs)~\cite{rbf} and Gaussian Processes (GPs)~\cite{williams2006gaussian} are popular choices when it comes to surrogate models.
Motivated by this, we devise a hybrid search algorithm for adapting EAs to low budget optimization problems, like HPO, by balancing the randomness of EA search moves with strategically generated search points via informative surrogates. As detailed next, the proposed \algo~uses a mix of surrogate models (RBF and GP) and mutation mechanism (based on EA) for solving HPO problems.

\section{The \algo~algorithm}
\label{sec:algo}
Our algorithm \textsc{S}urrogate-assis\textsc{t}ed Bay\textsc{e}si\textsc{a}n \textsc{D}ifferential \textsc{E}volution (\algo) starts with a stochastic RBF method~\cite{srbf,srbf_expensive}, which is parallel surrogate optimization algorithm, to estimate the functional landscape and then switches to an EA-based scheme guided by Bayesian Optimization (BO)~\cite{brochu2010tutorial}.
The design rationale is to perform initial exploration of the parameter space using the RBF model, and then transfer the knowledge to a GP-based Bayesian optimization framework, which is further augmented by a Differential Evolution (DE)  algorithm~\cite{differentialevolution}. DE is a population-based real-parameter global optimization technique that has gained widespread popularity as it is fast and reliable,  easy to implement, highly flexible, and robust to simple transformations~\cite{desurvey}. Recent work has shown the advantage of information sharing in DE framework specially for solving multimodal optimization~\cite{biswas2014inducing}.

\algo~begins by instantiating an experimental design like symmetric Latin hypercube design (SLHD)\footnote{We use SLHD as it allows an arbitrary number of design points/trial solutions and works well in practice~\cite{rbfopt}} to generate the design points (trial solutions) and evaluate them. The RBF-based surrogate model $\mathcal{R}$ is then used to approximate the objective function~\cite{srbf}. 
After building the model, we solve an auxiliary problem by optimizing the acquisition function to generate trial solution(s) for evaluation. In an iterative manner, we evaluate the trial solution(s) and update the surrogate model till $\lambda$ iterations. Then we switch to DE-based search for generating trial solution(s) till a termination criterion is met.

DE maintains a population $\mathbf{X}$ of $Np$ randomly generated $D$-dimensional real-parameter vectors representing trial solutions to a problem. We can represent the $i$\textsuperscript{th} solution (also called target vector), $i \in \left[1, 2, \ldots, Np \right]$, of the
population at the current generation $G$, $G=0,1,\ldots,G_{\max}$, as
\[\Vec{X} \u {i,G} = \left[x \u {1,i,G}, x \u {2,i,G}, x \u {3,i,G}, \ldots, x \u {D,i,G}\right].\]
Following the initialization, the solutions are improved iteratively through a series of steps$-$\textit{mutation}, \textit{crossover} or recombination, and \textit{selection}$-$till a termination criterion is met. For an in-depth understanding of the DE algorithm, kindly refer to~\cite{desurvey}.
In our approach, we propose a \textit{Bayesian mutation} mechanism to evolve new solutions (also called donor vectors) as follows.
\begin{equation}
    \Vec{V} \u {i, G} =
    \Vec{B} \u {r^i _1, G} + 
    F \cdot \left( \Vec{R} \u {r^i _2, G} - \Vec{R} \u {r^i _3, G} \right) + 
    \frac{1}{G} \cdot rand\left(1, D\right) \cdot \left( \Vec{X} \u {r^i _2, G} - \Vec{X} \u {r^i _3, G} \right),
    \label{eq:bayesmut}
\end{equation}
where 
the indices $r^i _1$, $r^i _2$, and $r^i _3$ are mutually exclusive integers randomly chosen from the range $\left[1, Np\right]$ such that $r^i _1 \neq r^i _2 \neq r^i _3 \neq i$, 
$rand\left(1, D\right)$ is a $D$-dimensional vector of uniformly distributed random numbers lying between $0$ and $1$,
$F$ is a scaling factor that typically lies in the interval $\left[0.4, 1\right]$,
$\Vec{B}$ and $\Vec{R}$ are the populations of trial solution(s) generated by a BO-based model and the RBF-based method, respectively.
The Bayesian mutation leverages the sample-efficiency of BO to search for solutions at promising regions of the parameter-space. The second component in Eq. (\ref{eq:bayesmut}) acts as a global search (exploratory) component while the third component gradually transitions from a global to local search as $G$ increases (and $1/G$ shrinks accordingly).

Next, the trial vectors are generated via the crossover operation in which every donor vector exchanges its components/features with its corresponding target vector.
DE can use two types of crossover$-$ \textit{binomial} and \textit{exponential}. Due to its flexibility, we adopt the binomial crossover as  
\begin{equation}
u_{j,i,G} =
    \left\{
        \begin{matrix}
            v_{j,i,G} & \text{if}\;( rand_{i,j}[ 0,1] \leq Cr\ or\ j==j_{rand})\\
            x_{j,i,G} & \text{otherwise} \hfill
        \end{matrix}
    \right.,
    \label{eq:binomial}
\end{equation}
where the crossover rate $Cr$ approximates the probability of recombination,
$rand\u {i,j}[0, 1]$ is an uniform random number generated for every $(i, j)$ pair,
$j\u{rand} \in [1, \ldots, D]$ is a randomly chosen index to ensure that $\Vec{U} \u {i,G}$ has at least one component from $\Vec{V} \u {i,G}$. Binomial crossover ensures that the number of components inherited by the trial vector from the donor vector roughly follows a binomial distribution.
Finally, a fitness-based selection mechanism takes place between the trial vector and the mutant vector to select the fitter one.
The pseudocode of the \algo~is outlined in Algorithm~\ref{algorithm}.

We use the \textsf{DE/rand/2/bin} model where two scaled difference vectors are used to mutate the target vector (generated by BO). We set the scaling factor $F$ and crossover rate $Cr$ to 0.7, $\lambda$ to 10 for activating the Bayesian mutation, and use population size $Np$ of 16 as per the competition guidelines.

\begin{algorithm}[!htp]
\DontPrintSemicolon
\caption{\algo~algorithm}
\SetKwInOut{Input}{Input}
\SetKwInOut{Output}{return}
\SetKwRepeat{Do}{do}{while}
\footnotesize
\Input{Objective function $f$, Parameter space $\mathbf{\Lambda}$, Population size $Np$, Evaluation budget $G \u {\max}$}
\KwResult{Best possible solution to $f$}
\Begin{
    Select an experimental design to instantiate trial solutions $\mathbf{X} \u {0} \in \mathbf{\Lambda}$ and evaluate them $f( \mathbf{X} \u {0} )$\;
    Use $f( \mathbf{X} \u {0} )$ to build a surrogate model 
    based on RBF\;
    \tcp*{Iterative improvement of the solutions till a termination criterion is met}
    \For{$G = 1, 2, \ldots, G \u{\max}$}
    {
        \eIf{$ G < \lambda $}
            {
                Generate trial solutions $\Vec{U} \u {i,G}, \,i \in \left[1, \ldots, Np \right]$ using the RBF-based method\;
            }
            {
                Generate solutions $\Vec{\mathcal{B}}$ and $\Vec{\mathcal{R}}$ from the RBF-based and BO-based model respectively\;
                Perform Bayesian mutation (\ref{eq:bayesmut}) to generate donor vectors/solutions\; 
                Simulate binomial crossover (\ref{eq:binomial}) to produce trial solutions  $\Vec{U} \u {i,G}$\;
            }
            Perform fitness-based selection to update the trial solutions as
            \begin{equation*}
            \Vec{X} \u {i,G+1} =
                \left\{
                    \begin{matrix}
                        \Vec{U} \u {i,G} & \text{if}\;f(\Vec{U} \u {i,G}) \leq f(\Vec{X} \u {i,G})\\
                        \Vec{X} \u {i,G} & \text{otherwise} \hfill
                    \end{matrix}
                \right.,
            \end{equation*}
            Use the trial solutions $\mathbf{X} \u {G+1}$ to update the surrogate models. 
            }
}
    Output the best solution $\Vec{X} \u {*} \in \mathbf{X} \u {G_{\max}}$\;

\label{algorithm}
\end{algorithm}
\normalsize

\subsection{Implementation details}
Our \algo~is built on top of the Bayesmark\footnote{Available in GitHub at \href{https://github.com/uber/bayesmark}{\textcolor{blue}{https://github.com/uber/bayesmark}}.} package, and uses its in-built \textsf{space} functionality to translate the parameter-space into a continuous search space using the idea of warping~\cite{warping}. 

The python Surrogate Optimization Toolbox (\pysot) is a  library of surrogate optimization techniques leveraging the Plumbing for Optimization with Asynchronous Parallelism (\poap) framework. For more details about these frameworks, kindly refer to~\cite{eriksson2019pysot}. Our code is based on \pysot~v0.3.3 and \poap~v0.1.26.
 We use a SLHD experimental design, a stochastic RBF surrogate with a cubic kernel and linear tail~\cite{srbf}, and the DYCORS strategy~\cite{srbf_expensive} for generating trial solutions, i.e., $\Vec{\mathcal{R}}$.

The other surrogate model is GP-based and is implemented using  \botorch, 
a recent programming framework that includes advanced Bayesian optimization techniques~\cite{botorch}. Since GPs involve expensive computation, we use batch BO to reduce the computation time. In batch BO, the design points are selected in parallel by doing joint optimization over the multiple design points. To do that we use the parallel counterpart of the sequential acquisition functions (EI, PI, UCB)~\cite{maximizing}. In our implementation, we use \textsf{qEI} acquisition function~\cite{ego} with randomized quasi-Monte Carlo (RQMC) sampling based on scrambled Sobol sequences for sample average approximation~\cite{caflisch1998monte,owen2003quasi}.
 The code of \algo~is made available at \textcolor{blue}{\tt https://github.com/subhodipbiswas/BayesianEvolution}. 

\section{Experimentation}
\label{sec:experiment}

\subsection{Methodology}
The \algo~model was developed for the blackbox optimization challenge\footnote{The starter kit and implementation details are available at \href{https://github.com/rdturnermtl/bbo_challenge_starter_kit/}{\textcolor{blue}{github.com/rdturnermtl/bbo\_challenge\_starter\_kit}}.} at NeurIPS 2020 under the team name \texttt{Better call Bayes}. The experimental setup consists of ML hyperparameter tuning problems. 
The competition guidelines allowed each optimizer to run for $16$ iterations using a batch size of $8$ for making suggestions on each benchmark problem. There was a strict time limit of 640 $s$ (or 40 $s$/iteration) which was not to be exceeded.
 Optimizers exceeding the time limits were cut off from making further suggestions and the best optima found till then was used.
We simulated $15$ runs\footnote{Each run consists of testing $9$ ML models on $6$ datasets using $2$ metric, thereby resulting in $108$ studies.} of \algo~on the benchmark resulting in $1620$ independent studies. The \textsf{Bayesmark} library was used to compute the performance of the optimizer on different benchmark problems.

\begin{figure}[!htp]
    \centering
    \begin{subfigure}[b]{0.48\linewidth}
        \includegraphics[width=0.95\linewidth,keepaspectratio]{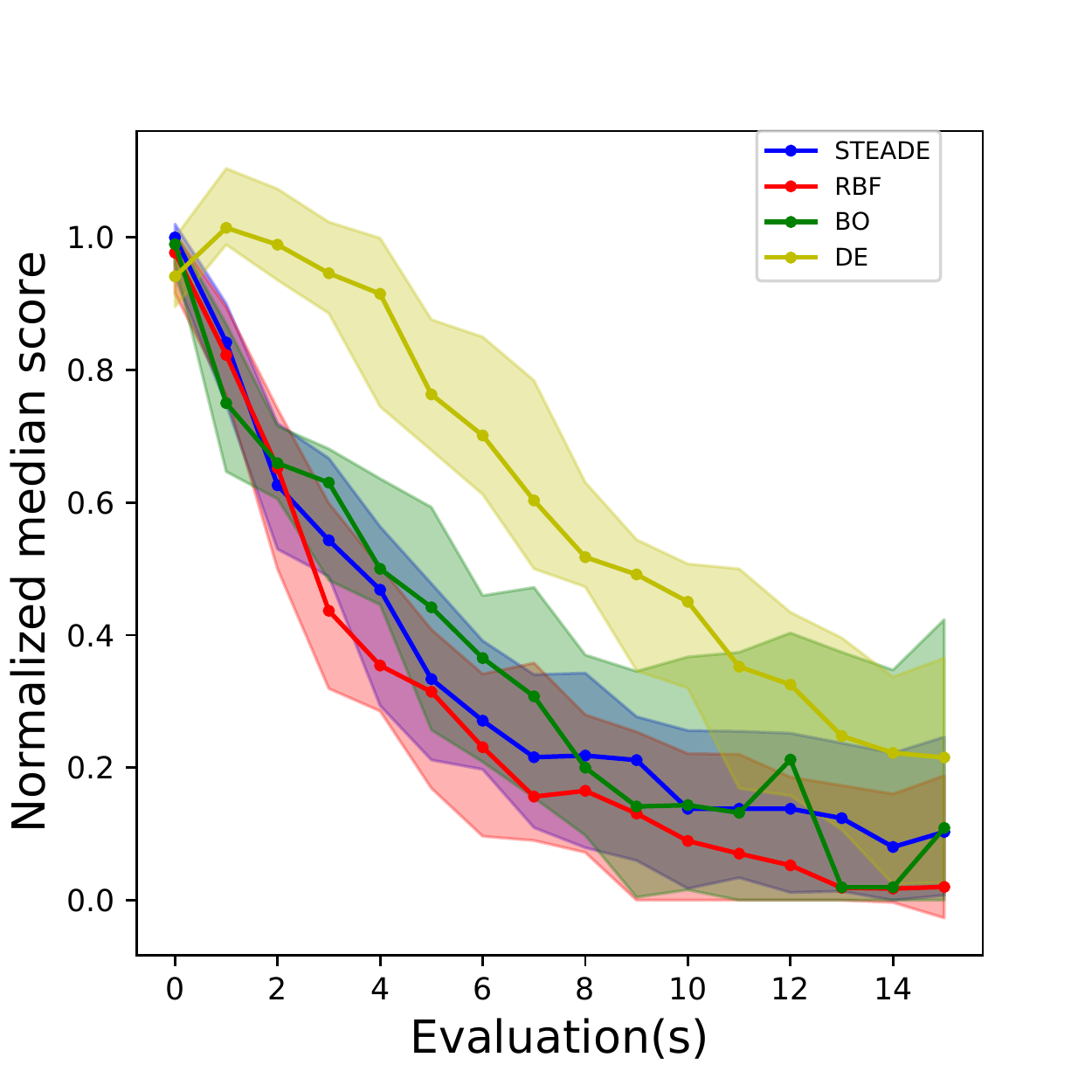}
        \label{fig:median}
    \end{subfigure}
~
   \begin{subfigure}[b]{0.48\linewidth}
        \includegraphics[width=0.95\linewidth,keepaspectratio]{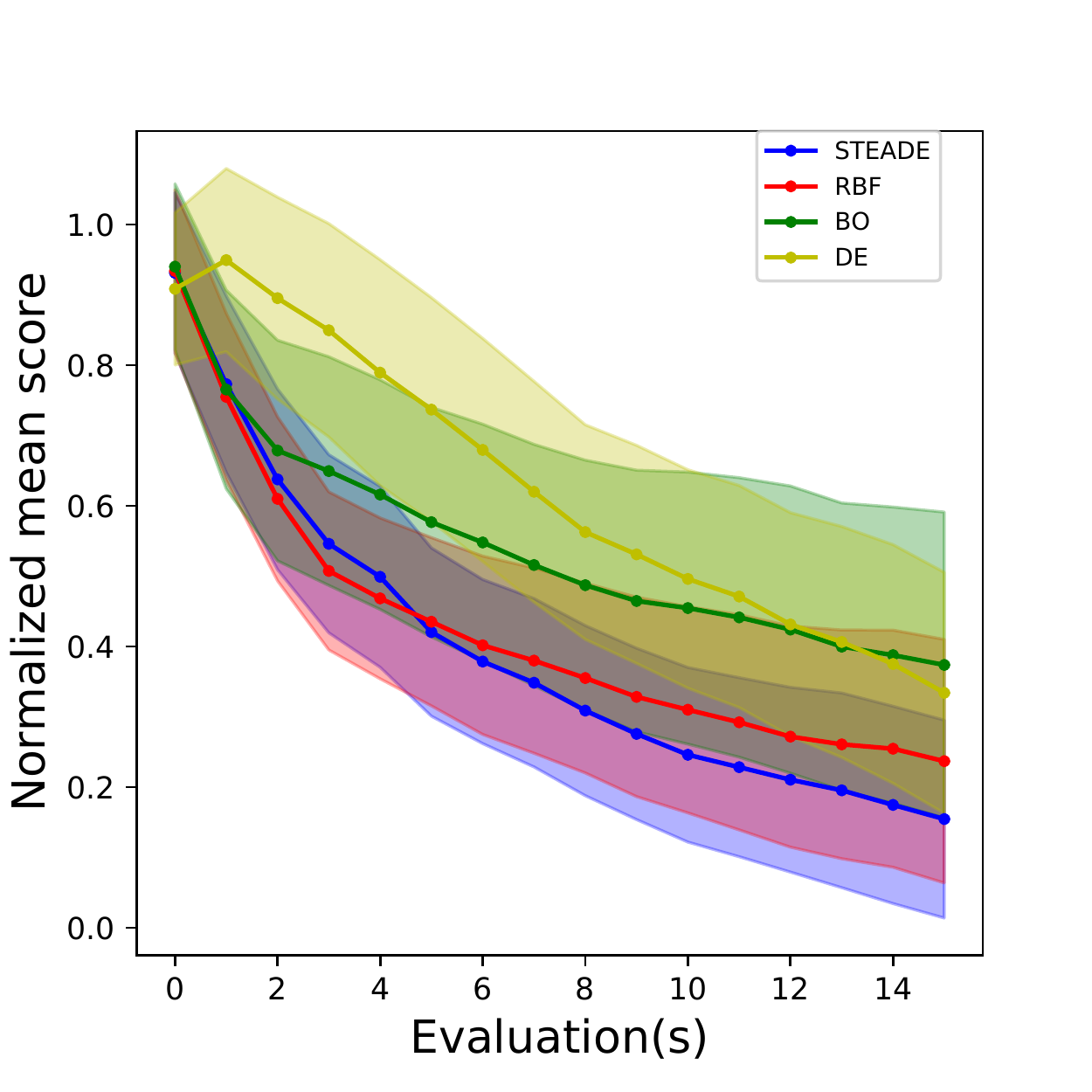}
        \label{fig:mean}
    \end{subfigure}
 \caption{The median and mean \texttt{visible\_to\_opt} score obtained by the different algorithms.}
    \label{fig:operators}
\end{figure}

Since \algo~is an composed of three components (RBF, BO and DE), we do ablation testing by selectively simulating each component under identical experimental conditions. The \texttt{visible\_to\_opt}, \texttt{generalization} and leaderboard\footnote{This score is based on the simulations performed in the local machine, and is different from our actual leaderboard score of $89.846$.} scores are reported in Table~\ref{tab:comparison}.
The \texttt{visible\_to\_opt} is the score seen by an optimizer (e.g, in a ML hyper-parameter tuning problem this can be the cross-validation error), whereas \texttt{generalization} is a related metric the optimizer does not get to see (e.g, error on held-out test set in a ML hyper-parameter tuning context).
For more details on how the mean, median and normalized mean are calculated, kindly refer to the documentation\footnote{See \href{https://bayesmark.readthedocs.io/en/latest/scoring.html\#analyze-and-summarize-results}{\textcolor{blue}{bayesmark.readthedocs.io/en/latest/scoring.html\#analyze-and-summarize-results}} for scoring details.}. We also plot the convergence profile of the different optimizers in Figure \ref{fig:operators} based on the   \texttt{visible\_to\_opt} score.

\begin{table*}[!h]
\centering
\caption{Comparative analysis of \algo~with its component algorithms.}
 \setlength{\tabcolsep}{3pt}
{\fontsize{9.0pt}{10.0pt}\selectfont
\begin{tabular}{@{}c|ccc|ccc|c@{}}
\toprule
\multirow{2}{*}{
Models
}  & 
\multicolumn{3}{c|}{\texttt{visible\_to\_opt}} &
\multicolumn{3}{c|}{\texttt{generalization}} &
\multirow{2}{*}{\texttt{Leaderboard}}
\\ \cline{2-7} &
    Median      &    Mean    &   Normalized Mean   &
    Median      &    Mean    &   Normalized Mean   &
    
    \\ 
    \midrule
BO 
&
    0.10906  & 0.03151   &  0.37376  &
    2.36782  & 0.29934   & 2.58823  &
96.8487
    \\ 
DE    
&
     0.21536  &  0.02817  &  0.33408 &  
     2.5  &  0.29478  & \textbf{2.54886} &  
     97.1833
     \\ 
RBF
&        
    0.02002  &  0.01998  &  0.23693  &
    2.62130  &  0.30189 & 2.61027  &
    98.0024
     \\ 
STEADE
&        
   0.10301  &  0.01304  &  \textbf{0.15463} &
   2.51204  &  0.297115 & 2.56901 &
   \textbf{98.6963}
  \\ 
  \bottomrule
\end{tabular}
\par}
\label{tab:comparison}
\end{table*}

\subsection{Results and Discussions}
In Figure~\ref{fig:operators}, we notice that the random nature of DE search move results in low search efficiency during the initial stages of run while the surrogate-based models perform informed search due to their functional estimation capability. The search efficiency of the surrogate-based models gradually fall and after a certain point, there seems to be a saturation in a model's knowledge about the functional landscape. In such scenarios, one plausible way is to switch to a different model for carrying out further search. In \algo, we transfer the knowledge from an RBF interpolation to a GP-based model. The solutions generated by the GP are further perturbed using a DE-based mutation. The efficacy of our approach is evident from Table~\ref{tab:comparison}, wherein \algo~is able to achieve better normalized mean score and leaderboard score in comparison to the baselines.

Even though DE has been extensively applied to solve cheap optimization problems, interestingly, very few works have explored its applicability to \textit{low-budget} (computationally expensive) optimization problems.
In this work, we take a step towards this direction and demonstrate how EAs can benefit from surrogate models in such scenarios. However, the increased cost of surrogates somewhat limits the applicability of such hybrid models to cheap/moderate budget optimization problems. In order to make that possible, we need to design careful learning mechanisms that will transition from the expensive surrogate-based model to a cheap EA model depending on the perceived gain in functional value with respect to computational time. This is a possible research direction that we wish to undertake in near future.

\small

\bibliographystyle{plain}
\bibliography{references}

\end{document}